\pdfoutput=1

\documentclass[11pt]{article}

\usepackage{ACL2023}

\usepackage{times}
\usepackage{latexsym}

\usepackage[T1]{fontenc}

\usepackage[utf8]{inputenc}

\usepackage{microtype}

\usepackage{inconsolata}

\usepackage{graphicx}
\usepackage{hyperref}
\usepackage{booktabs}
\usepackage{multirow} 
\usepackage{enumitem}
\usepackage{url}
\usepackage{wrapfig}
\usepackage{listings}
\usepackage{subcaption}

\usepackage[capitalize,noabbrev]{cleveref}

\newcommand{\besto}[1]{\textcolor{violet}{#1}}
\newcommand{\bests}[1]{\textbf{#1}}

\newsavebox{\mybox}

\title{Examining the Emergence of Deductive Reasoning in Generative Language Models}

\author{Peter Belcak \and Luca A. Lanzend\"orfer  \and Roger Wattenhofer \\
    ETH Z\"urich \\
  \texttt{\{belcak,lanzendoerfer,wattenhofer\}@ethz.ch}
}

\begin{document}

\maketitle

\begin{abstract}
We conduct a preliminary inquiry into the ability of generative transformer models to deductively reason from premises provided.
We observe notable differences in the performance of models coming from different training setups and find that the deductive reasoning ability increases with scale.
Further, we discover that the performance generally does not decrease with the length of the deductive chain needed to reach the conclusion, with the exception of OpenAI GPT-3 and GPT-3.5 models.
Our study considers a wide variety of transformer-decoder models, ranging from 117 million to 175 billion parameters in size.
\end{abstract}

\section{Introduction}
\label{section:introduction}
It is inherently desirable for foundation language models to be capable of reasoning.
This is especially true for generative models, where the hope is that the models could ultimately hold a profound conversation or continue a chain logical reasoning with human-like performance.

The most tractable of modes of reasoning is deduction, which employs premises and general laws to reach conclusions.
In contrast to inductive reasoning, which deals with generalisation of observed phenomena, and abductive reasoning, which is invoked to find a likely cause for a conclusion in some context of knowledge, deduction is the mode of reasoning used to find the consequences of available knowledge.
As such, it is a natural and necessary part of any generative language model -- a model trained on a large corpus of knowledge and queried to make use of it must internally perform a chain of deductions to produce a true consequence of its knowledge as prompted.

Our aim is to study large language models, assess their ability to perform deductive reasoning, and see how they fare as the size of the model and the complexity of required deductions grow.
Note that the use of deductive reasoning is implicitly assumed when interacting with a large generative language model.
It is expected that regardless of the prompt, the model will produce an output that logically follows either from the model's knowledge or the context provided, and that this output complies with the prompting instruction.
Even when asked to perform creative writing or continue a paragraph, it is expected that the generated text will logically follow or at least not contradict the context \citep{yuan2022wordcraft}.
Hence, our study considers the native scenario in which no particular prompting instruction is provided.

Large language models have been shown to excel at memorising fragments of factual statements in their training data and at internalising highly co-occurring factual associations \citep{li2022pre,carlini2022quantifying}.
A study that would hope to evaluate model's deductive reasoning ability while requiring ``common sense'' \citep{lourie2021unicorn,singh2021com2sense} would thus also be indirectly evaluating its common sense knowledge.
We therefore steer clear of any factual knowledge and carry out our evaluation under a restricted setting, in which we provide the model with a context of premises (a theory) and ask a single \textit{yes/no} question, insisting that
\begin{enumerate}[label=P\arabic*.]
    \item \label{item:P1} all premises are generic in their wording (they only involve neutral entities and properties),
    \item \label{item:P2} all premises are precise in their meaning (there are no ambiguities arising from synonymy, paraphrasing, or categorical adherence),
    \item \label{item:P3} all conclusions that may be queried by the question can be arrived at by (possibly several) applications of \textit{modus ponens} (if \texttt{A} and \texttt{if A then B} then \texttt{B}), and that 
    \item \label{item:P4} all premises that hold true are explicitly listed and provided to the model, with any fact that is not a premise nor a consequence assumed not to be true (i.e. we make the closed-world assumption).
\end{enumerate}

We set out to answer three research questions:
\begin{enumerate}[label=Q\arabic*.]
    \item \label{item:reasoningAgainstSize} How does the deductive reasoning ability (DRA) of large language models evolve with their growing size?
    \item \label{item:reasoningAgainstDepth} How does the DRA of general-purpose generative large language models depend on the number of deductive steps required to reach the conclusions?
    \item \label{item:reasoningAgainstType} Do the specifics of a training setup, such as the model variant, dataset multilingualism, or dataset size have a decisive influence on the DRA?
\end{enumerate}
We measure DRA by binary accuracy on the answers of the model; in other words, by the proportion of the instances in which the model answered the question about the theory given as context correctly.
A detailed account of our experimental setup is presented in \Cref{section:experiments}.

\hyperref[item:reasoningAgainstSize]{Q1} sets the first dimension of our inquiry -- the impact of the size of the language model on DRA.
Even relatively small pre-trained language models (BERT, RoBERTa) have been shown to perform well on isolated instances of unstructured natural language inference when fine-tuned to that end \citep{wang2021entailment,sun2020self}.
Models of similar size were also fine-tuned for small structured languages to determine whether a statement follows from a theory, but it was observed that the performance of such models deteriorated rapidly when the number of deductive steps (``depth'') required to arrive at the answer in test examples was higher than what had been seen in training \citep{clark2020transformers}.
Instances of models twice as large were successfully fine-tuned for proof generation in structured language, though it should be noted that when asked to generate the proof in a single run, the models suffered from the same limitation on the question depth \citep{tafjord2020proofwriter}.
The largest of language models remain largely untested in this regard, though it has been shown that they perform rather poorly even on elementary reasoning and arithmetic tasks unless prompted through chains of thought or by explicitly providing the relevant algorithm \citep{wei2022chain,zhou2022teaching}.
To answer \hyperref[item:reasoningAgainstSize]{Q1}, we comprehensively evaluate 16 models in total, ranging from 117 million to 175 billion parameters in size.

\hyperref[item:reasoningAgainstDepth]{Q2} asks about the influence of the reasoning depth required on model DRA.
As already mentioned, previous work \citep{clark2020transformers,tafjord2020proofwriter} has shown that even the deductive performance of models \textit{fine-tuned} in the context of deductive reasoning considerably decreases once the depth required for inference climbs above the maximum depth seen in training.
We test all 16 models considered for \hyperref[item:reasoningAgainstSize]{Q1} for depths $0$, $1$, $2$, and $3$, paying attention to any changes in scores.

\hyperref[item:reasoningAgainstType]{Q3} leverages the experiments conducted for \hyperref[item:reasoningAgainstSize]{Q1}-\hyperref[item:reasoningAgainstDepth]{2} in an attempt to see whether there might be other factors influencing DRAs of foundation language models.
While large language models tend to be trained on enormously large datasets, they vary in many details of their training, for example the proportions of different languages used, the use of code versus the pure use of the natural language, and the level to which edited sources (e.g. books, news) are emphasised over spontaneous language (e.g. chats, tweets).
We inspect the results of all experiments for patterns aligned with the boundaries between different training setups.

In \Cref{section:results}, we answer each of the questions \hyperref[item:reasoningAgainstType]{Q1}-\hyperref[item:reasoningAgainstDepth]{3} in turn and comment on our further findings.
We structured our study under the hypothesis that the model size would play a decisive role in model DRA, that the model performance would quickly decrease with depth, and that the role of the training setup would be negligible.

\textsc{Reproducibility statement.}
For full reproducibility, we make the datasets, runtime configurations, and all our code available at \textit{anonymised}.

\section{Experiments}
\label{section:experiments}

\subsection{Data}
We use the closed-world variant of the RuleTakers dataset introduced in \cite{clark2020transformers} and further expanded in \cite{tafjord2020proofwriter}.
The dataset contains groups of 100,000 randomly generated theories for depths ranging from $0$ to $5$, written in structured language.
A group of depth $k$ contains yes/no questions that ask about statements (dis)provable in up to $k$ deductive steps, generated at random but such that the dataset is balanced on average.
The dataset further contains two special groups, \texttt{NatLang} and \texttt{Birds-Electricity}, each containing examples written freely in natural language and with separating punctuation where appropriate, but still abiding by the principles \hyperref[item:P1]{P1}-\hyperref[item:P4]{4}.
\Cref{appendix:input_examples} gives examples of theories written both in structured and natural language.

\subsection{Inputs}
\label{section:inputs}
Inputs given to our models consist of the theory, leading examples (if considered), and the question that is to be answered.
The three components are concatenated, with whitespace and punctuation inserted where appropriate.
The leading examples are provided in the form ``Question? Answer.'', where question is as provided by the dataset and answer is either ``Yes'' or ``No''.

There are always either $0$ or $3$ leading examples provided.
In the latter case, we give one leading example for each answer, determined at random, and a distinct example with either answer, also determined at random.
This was done to make sure that the models (especially the ones below 3bn parameters) are not tempted by continuing a series of three ``Yes'' or three ``No'' answers instead of answering the question.

\subsection{Evaluation}
\label{section:evaluation}
We ran the inferences of 11 open-source models on the test slice of the dataset locally, and further evaluated 4 recent OpenAI models via the OpenAI API.
In either case, we provided the model with inputs as in \Cref{section:inputs} and asked it to make predictions for the next token.

\subsubsection{Open-source models}
The open-source models had between 117 million and 66 billion parameters and can be sorted into three families based on their training setup: OpenAI GPT-1/2 models, Bloom models, and Facebook OPT models.
The smaller from among those models would answer our questions with ``Yes'', ``No'', but also with `` True'', ``\texttt{<line-break>}False'', and with varying capitalisation.
Further, ``Yes'' and ``No'' were not always the most favoured answers, and the models sometimes exhibited a tendency to either continue the theory by inventing more facts, or continue asking questions without answering the question they were asked.
To evaluate the models consistently, we therefore inspected the predicted probability distribution of the first token and considered the model to have answered ``Yes'' if it answered ``Yes'' or ``True'' irrespective of capitalisation, and similarly considered it to have answered ``No'' if it answered with ``No'' or ``False''.

We also experimented with equipping some of the $\leq 1$bn-parameter models with a linear classification head on the embeddings of the final token as in \cite{radford2018improving} and training it to perform yes/no classification with the base either completely model frozen or frozen except for the final transformer layer.
We did not observe any improvements over the results of the evaluation as above.
Due to us wishing to evaluate the original version of the model and because of the limits on our computational resources, we did not attempt to fine-tune entire networks.

\subsubsection{OpenAI GPT-3 and GPT-3.5 models}
Since the OpenAI API does not provide access to the probabilities of the tokens predicted, we configured the models with temperature 1 and asked them to predict a single token.
In all instances where leading examples were provided, the models were consistently outputting either ``Yes'' or ``No'' with some variance in capitalisation.
In instances where no leading examples were given, all models but the GPT-3.5 ``da Vinci'' tended to behave similarly to $\leq 3$bn-parameter open-source models and either continued the theory or continued generating questions instead of answering.
Due to the lack of information about other predicted tokens we did not perform the full set of experiments with no leading examples.

\begin{table*}[t!]
\centering
\scalebox{0.85}{
    \begin{tabular}{l|c|rrrrr|rrrr|rr}
    
    \toprule
    Data & LEs & \multicolumn{11}{c}{Model}\\
    \midrule
    
    
    \multicolumn{2}{c}{} &
    \parbox[t]{10mm}{\multirow{4}{*}{\centering\rotatebox[origin=c]{70}{GPT-1}}} &
    \parbox[t]{10mm}{\multirow{4}{*}{\centering\rotatebox[origin=c]{70}{GPT-2 small}}} &
    \parbox[t]{10mm}{\multirow{4}{*}{\centering\rotatebox[origin=c]{70}{GPT-2 medium}}} &
    \parbox[t]{10mm}{\multirow{4}{*}{\centering\rotatebox[origin=c]{70}{GPT-2 large}}} &
    \parbox[t]{10mm}{\multirow{4}{*}{\centering\rotatebox[origin=c]{70}{GPT-2 XL}}} &
    \parbox[t]{10mm}{\multirow{4}{*}{\centering\rotatebox[origin=c]{70}{Bloom 560m}}} &
    \parbox[t]{10mm}{\multirow{4}{*}{\centering\rotatebox[origin=c]{70}{Bloom 1b1}}} &
    \parbox[t]{10mm}{\multirow{4}{*}{\centering\rotatebox[origin=c]{70}{Bloom 3b}}} &
    \parbox[t]{10mm}{\multirow{4}{*}{\centering\rotatebox[origin=c]{70}{Bloom 7b1}}} &
    \parbox[t]{10mm}{\multirow{4}{*}{\centering\rotatebox[origin=c]{70}{OPT 30b}}} &
    \parbox[t]{10mm}{\multirow{4}{*}{\centering\rotatebox[origin=c]{70}{OPT 66b}}} \\
    \multicolumn{9}{c}{} \\
    \multicolumn{9}{c}{} \\
    \multicolumn{9}{c}{} \\
    \multicolumn{9}{c}{} \\
    \multicolumn{9}{c}{} \\
    
    \midrule
    \multicolumn{2}{l}{} & \textit{117mn} & \textit{124mn} & \textit{355mn} & \textit{774mn} & \textit{1.5bn} & \textit{560m} & \textit{1.1bn} & \textit{3bn} & \textit{7.1bn} & \textit{30bn} & \textit{66bn} \\
    \midrule
        \multirow{2}{*}{\texttt{Depth 0}}                   & none   & +0.08  & \bests{+2.15}    & +1.89  & +1.33  & -1.20  & \bests{\besto{+6.76}}  & +1.05  & +3.13  & +5.64  & +1.68   & \bests{+4.20}  \\
                                                            & YNR    & \bests{-0.05}  & -24.32   & -19.79 & -11.14 & -14.77 & -0.98  & -5.12  & \bests{\besto{+3.97}}  & -0.43  & -11.47  & \bests{-7.14}  \\
        \cmidrule(r){3-13}                                                                                                                                                 
        \multirow{2}{*}{\texttt{Depth 1}}                   & none   & -0.09  & \bests{+0.83}    & +0.13  & +0.35  & -1.24  & \bests{\besto{+3.57}}  & +1.11  & +1.19  & +3.12  & +0.39   & \bests{+3.11}  \\
                                                            & YNR    & \bests{-0.01}  & -7.52    & -5.79  & -2.35  & -3.50  & +0.07  & -0.14  & \besto{\bests{+0.94}}  & +0.29  & -3.00   & \bests{-2.54}  \\
        \cmidrule(r){3-13}                                                                                                                                                 
        \multirow{2}{*}{\texttt{Depth 2}}                   & none   & -0.03  & \bests{+2.59}    & +2.22  & +0.69  & -1.17  & \bests{\besto{+5.95}}  & +1.78  & +3.57  & +4.56  & +0.78   & \bests{+2.68}  \\
                                                            & YNR    & -0.04  & -3.58    & -1.80  & -0.57  & \bests{+0.44}  & +1.20  & +1.83  & \bests{+2.29}  & +1.29  & +0.80   & \bests{\besto{+2.68}}  \\
        \cmidrule(r){3-13}                                                                                                                                                 
        \multirow{2}{*}{\texttt{Depth 3}}                   & none   & +0.09 & +3.43  & \bests{+3.43}  & +0.55  & -0.86  & \bests{\besto{+6.57}}  & +2.32  & +4.22  & +5.24  & +1.18   & \bests{+4.88}  \\
                                                            & YNR    & +0.01  & -1.99    & -0.08  & +1.53  & \bests{+2.12}  & +1.36  & +2.53  & \bests{+2.72}  & +1.41  & +2.06   & \bests{\besto{+4.65}}  \\
    \bottomrule 
    
    \end{tabular}
}
\caption{
    Results of our evaluation of open-source models.
    The models are listed by family and then given in the order of their growing size.
    ``LEs'' denotes leading examples, where ``none'' means no leading examples were provided and ``YNR'' means that examples of ``Yes'', ``No'', and a random question were given.
    The metric used is binary accuracy displayed in percentage points relative to the baseline coin-flip performance of 50\%, measuring the proportion of answers that were correct. \besto{Emphasis} and \bests{emphasis} mark the best performances overall and for the model family.
}
\label{table:results_opensource_structlang}
\end{table*}

\section{Results}
\label{section:results}

\begin{figure}
  \begin{center}
    \includegraphics[width=0.45\textwidth]{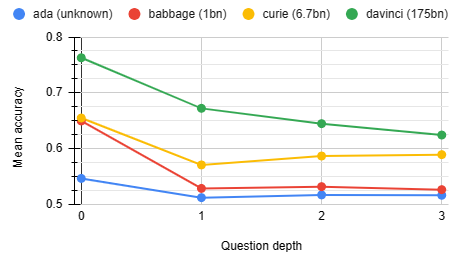}
    \caption{
        The performance of OpenAI GPT-3 and GPT-3.5 models plotted against the increasing length of questions in the dataset.
        The detailed results are in \Cref{appendix:results_openai}.
    }
    \label{figure:results_openai}
  \end{center}
\end{figure}

\Cref{table:results_opensource_structlang} lists the results of our experimentation on the open-source models, and \Cref{figure:results_openai} shows the performance of the OpenAI models available only through the OpenAI API.
Appendices \ref{appendix:results_nl}-\ref{appendix:results_openai} give the detailed results for natural language datasets and OpenAI GPT-3 and GPT-3.5 models.

We hypothesized that an increase in model size would correspond to an increase in the model's DRA.
\Cref{table:results_opensource_structlang} shows that this is often the case within respective model families except OpenAI GPT-1 and 2 models but not across the board -- for example, OPT-66b outperforms OPT-30b but underperforms Bloom-7b.

\textbf{\hyperref[item:reasoningAgainstSize]{Q1}: How does the deductive reasoning ability (DRA) of large language models evolve with their growing size?}
We find that with the training setup and model architecture kept constant, larger model size leads to higher DRA, but also that this effect can often be compensated by choices made in the training configuration.
Previous work has shown that the DRA of models fine-tuned for deduction decreases on data that requires longer chains of reasoning that have been seen in training.
\Cref{table:results_opensource_structlang} shows that this does not seem to be the case for foundation models as we evaluated them, that the performance of many models on deeper questions is similar to that on shallower questions, and that it sometimes even increases, especially for OpenAI GPT-1/2 models.

\textbf{\hyperref[item:reasoningAgainstDepth]{Q2}: How does the DRA of general-purpose generative large language models depend the number of deductive steps required to reach the conclusions?}
For depths considered in our study, it remains largely similar throughout, except for the OpenAI GPT-3 and 3.5 models, for whom the performance decreases as seen in \Cref{figure:results_openai}.

\textbf{\hyperref[item:reasoningAgainstType]{Q3}: Do the specifics of a training setup, such as the model variant, dataset multilingualism, or dataset size have a decisive influence on the DRA?}
Yes. As noted for \hyperref[item:reasoningAgainstSize]{Q1}, the differences between model training setups in many cases play a larger role than the size of the model.
An example outlier is Bloom-560m, which comfortably outperforms all larger Bloom models, and Bloom setting as a whole, which in turn outperforms much larger OPT and GPT-2 XL models.
This is perhaps due to the multi-lingual training setup (including multiple programming languages) of Bloom models, which is in contrast to natural language English used for GPT-1/2 and OPT models.

In summary, we conclude that all our initial hypotheses were partially incorrect, and that the subject of deductive reasoning ability in large language models warrants further examination from a wider set of angles.

\section{Limitations}
Our work only considers the token probabilities directly as provided by the generative models.
Since the models are trained with generation of patches of text in mind, it can be argued that the model might in fact be trying to output a verbose answer.
We attempt to to mitigate this by priming the model, but higher accuracy could perhaps be achieved by fine-tuning the language modelling head.

\bibliography{custom.bib}
\bibliographystyle{acl_natbib}

\appendix
\onecolumn

\newpage
\section{Input Examples}
\label{appendix:input_examples}
\begin{figure*}[h]
        \centering
\begin{lrbox}{\mybox}%
\begin{lstlisting}
Erin is kind. Erin is quiet. Erin is round. Fiona is green. Gary is big.
Gary is green. Harry is quiet. If Harry is round then Harry is young.
If something is kind then it is big. If something is green and not round 
then it is white. Quiet things are green. If Fiona is young and Fiona is
not round then Fiona is quiet. If something is white then it is young.
\end{lstlisting}
\end{lrbox}
        \scalebox{0.90}{\usebox{\mybox}}
        \caption{An example of a theory of the RuleTakers dataset, written in structured language. Whether Erin is round would be a question of depth $0$, whether Erin is big would be a question of depth $1$, and whether Harry is young would be a question of depth $3$.}
        \label{figure:theoryExample1}
\end{figure*}

\begin{figure*}[h]
        \centering
\begin{lrbox}{\mybox}%
\begin{lstlisting}
For being so cold, it's good Alan can remain nice. Charlie might be rough
and red but he's actually very kind. Dave can be rough and cold, but he is 
also green, an avid gardener. Fred is a round and rough around the edges, 
and he is also big. People who have green body paint and act kind to others 
are quite young. A nice, green, big person is also sure to be a red person. 
Cold people that are big and red are usually nice. Someone with rough and 
green feet is invariably kind. Tom is a rough, young person to know, he is 
very green at his job but he is very round from his weight.
\end{lstlisting}
\end{lrbox}
        \scalebox{0.89}{\usebox{\mybox}}
        \caption{An example of a theory from the \texttt{NatLang} part of the RuleTakers dataset, paraphrased into natural language. Here, ``Is Dave green?'' is a question of depth $0$, ``Is Dave kind?'' is a question of depth $1$, and `Is Dave young?'' is a question of depth $3$.}
        \label{figure:theoryExample2}
\end{figure*}

\begin{figure*}[h]
        \centering
\begin{lrbox}{\mybox}%
\begin{lstlisting}
Erin is kind. Erin is quiet. Erin is round. Fiona is green. Gary is big.
Gary is green. Harry is quiet. ... Erin is round? Yes. Harry is round? No.
Fiona is green? Yes. Fiona is quiet?
\end{lstlisting}
\end{lrbox}
        \scalebox{0.89}{\usebox{\mybox}}
        \caption{An instance of input based on the theory from \Cref{figure:theoryExample1}, supplemented with three leading examples of depth $0$ and the question to be answered by the model.}
        \label{figure:theoryExample3}
\end{figure*}

\newpage
\section{Detailed Results of the Evaluation on Natural Language Datasets}
\label{appendix:results_nl}

\begin{table*}[h!]
\centering
\scalebox{0.710}{
    \begin{tabular}{l|c|rrrrr|rrrr|rr}
    
    \toprule
    Data & LEs & \multicolumn{11}{c}{Model}\\
    \midrule
    
    
    \multicolumn{2}{c}{} &
    \parbox[t]{10mm}{\multirow{4}{*}{\centering\rotatebox[origin=c]{70}{GPT-1}}} &
    \parbox[t]{10mm}{\multirow{4}{*}{\centering\rotatebox[origin=c]{70}{GPT-2 small}}} &
    \parbox[t]{10mm}{\multirow{4}{*}{\centering\rotatebox[origin=c]{70}{GPT-2 medium}}} &
    \parbox[t]{10mm}{\multirow{4}{*}{\centering\rotatebox[origin=c]{70}{GPT-2 large}}} &
    \parbox[t]{10mm}{\multirow{4}{*}{\centering\rotatebox[origin=c]{70}{GPT-2 XL}}} &
    \parbox[t]{10mm}{\multirow{4}{*}{\centering\rotatebox[origin=c]{70}{Bloom 560m}}} &
    \parbox[t]{10mm}{\multirow{4}{*}{\centering\rotatebox[origin=c]{70}{Bloom 1b1}}} &
    \parbox[t]{10mm}{\multirow{4}{*}{\centering\rotatebox[origin=c]{70}{Bloom 3b}}} &
    \parbox[t]{10mm}{\multirow{4}{*}{\centering\rotatebox[origin=c]{70}{Bloom 7b1}}} &
    \parbox[t]{10mm}{\multirow{4}{*}{\centering\rotatebox[origin=c]{70}{OPT 30b}}} &
    \parbox[t]{10mm}{\multirow{4}{*}{\centering\rotatebox[origin=c]{70}{OPT 66b}}} \\
    \multicolumn{9}{c}{} \\
    \multicolumn{9}{c}{} \\
    \multicolumn{9}{c}{} \\
    \multicolumn{9}{c}{} \\
    \multicolumn{9}{c}{} \\
    
    \midrule
    \multicolumn{2}{l}{} & \textit{117mn} & \textit{124mn} & \textit{355mn} & \textit{774mn} & \textit{1.5bn} & \textit{560m} & \textit{1.1bn} & \textit{3bn} & \textit{7.1bn} & \textit{30bn} & \textit{66bn} \\
    \midrule
        \multirow{2}{*}{\texttt{NatLang}}                   & none   & -0.49 & \bests{\besto{+0.84}}  & -0.41  & +0.50   & -0.52 & 0.34  & -0.60  & -0.27  & \bests{0.55}  & -0.71   & \textit{N/A}  \\
                                                            & YNR    &  \bests{\besto{0.00}} & -3.37  & -1.89  & -2.51   & -1.97 & -0.51    & \bests{-0.50}  & -0.87  & -1.29  & -3.06  & \textit{N/A}  \\
        \cmidrule(r){3-13}                                                                                                                                                  
        \multirow{2}{*}{\texttt{Birds-Elec.}}               & none   & -0.27 &  0.00  & \bests{\besto{+6.36}}  & -12.01  & -0.04 & -5.52    & -6.41  & \bests{+1.19}  & -7.42  & -7.77   & \bests{+2.58}  \\
                                                            & YNR    &  \bests{0.00} & -2.49  & -6.81  & -6.53   & -1.04 & -7.59    & -9.37  & -4.72  & \bests{\besto{1.46}}  & \bests{-2.72}   & -6.80  \\
    \bottomrule 
    
    \end{tabular}
}
\caption{
    Results of our evaluation of open-source models on natural language datasets.
    The models are listed by family and then given in the order of their growing size.
    ``LEs'' denotes leading examples, where ``none'' means no leading examples were provided and ``YNR'' means that examples of ``Yes'', ``No'', and a random question were given.
    The metric used is binary accuracy displayed in percentage points relative to the baseline coin-flip performance of 50\%, measuring the proportion of answers that were correct.
    \besto{Emphasis} and \bests{emphasis} mark the best performances overall and for the model family.
}
\label{table:results_opensource_natlang}
\end{table*}

\newpage
\section{Detailed Results of the Evaluation of OpenAI GPT-3 and 3.5 models}
\label{appendix:results_openai}

\begin{table*}[h!]
\centering
\scalebox{1.0}{
    \begin{tabular}{l|c|rrrr}
    
    \toprule
    Data & LEs & \multicolumn{4}{c}{Model}\\
    \midrule
    
    
    \multicolumn{2}{c}{} &
    GPT-3 Ada &
    GPT-3 Babbage &
    GPT-3 Curie &
    GPT-3.5 da Vinci \\
    
    \midrule
    \multicolumn{2}{l}{} & \textit{unknown} & \textit{1bn} & \textit{6.7bn} & \textit{175bn} \\
    \midrule
        \multirow{1}{*}{\texttt{Depth 0}}                    
                                                            & YNR    & +4.67  & +14.96 & +15.49 & \bests{\besto{+26.28}} \\
                                                                                                                                                       
        \multirow{1}{*}{\texttt{Depth 1}}                   
                                                            & YNR    & +1.23  & +2.88  & +7.09  & \bests{\besto{+17.23}} \\

        \multirow{1}{*}{\texttt{Depth 2}}                   
                                                            & YNR    & +1.72  & +3.20  & +8.68  & \bests{\besto{+14.48}} \\

        \multirow{1}{*}{\texttt{Depth 3}}                   
                                                            & YNR    & +1.65  & +2.66  & +8.93  & \bests{\besto{+12.44}} \\
    \bottomrule 
    
    \end{tabular}
}
\caption{
    Results of our evaluation of OpenAI GPT-3 and GPT-3.5 models on structured langauge datasets.
    These were also the results used in \Cref{figure:results_openai}.
    The models are listed in the order of their growing size.
    ``LEs'' denotes leading examples, ``YNR'' means that examples of ``Yes'', ``No'', and a random question were given.
    As explained in \Cref{section:evaluation}, we did not comprehensively evaluate the models for their performance with no leading examples due to the limited information available through the OpenAI API.
    The metric used is binary accuracy displayed in percentage points relative to the baseline coin-flip performance of 50\%, measuring the proportion of answers that were correct.
    \besto{Emphasis} and \bests{emphasis} mark the best performances overall and for the model family.
}
\label{table:results_openai_structlang}
\end{table*}

\end{document}